\documentclass[11pt]{article}

\usepackage[margin=1in]{geometry}
\usepackage[T1]{fontenc}
\usepackage[utf8]{inputenc}
\usepackage{lmodern}
\usepackage{microtype}
\usepackage{graphicx}
\usepackage{booktabs}
\usepackage{multirow}
\usepackage[table]{xcolor}
\usepackage{pifont}
\usepackage{bm}
\usepackage{amsmath,amssymb}
\DeclareMathOperator*{\argmin}{arg\,min}
\usepackage{cite}
\usepackage[colorlinks=true,linkcolor=blue,citecolor=blue,urlcolor=blue]{hyperref}
\usepackage[capitalise,nameinlink]{cleveref}

\newcommand{\eg}{e.g.}

\hypersetup{
  pdftitle={ForensicsTok: Forensics-Guided Tokenized Modeling for Image Tampering Localization},
  pdfauthor={Lei Xu, Haowei Wang, Shen Chen, Taiping Yao, Bin Li, Changsheng Chen}
}

\title{ForensicsTok: Forensics-Guided Tokenized Modeling for Image Tampering Localization}

\author{%
Lei Xu$^{1}$ \quad
Haowei Wang$^{2}$ \quad
Shen Chen$^{2}$ \quad
Taiping Yao$^{2}$ \quad
Bin Li$^{1}$ \quad
Changsheng Chen$^{1}$\\[0.6em]
\small $^{1}$Shenzhen University \quad
\small $^{2}$Tencent Youtu Lab \quad
}

\date{}

\begin{document}
\maketitle

\begin{abstract}
Multi-modal Large Language Models (MLLMs) offer powerful reasoning for forensic tasks, yet existing approaches utilizing exogenous segmentation decoders often suffer from suboptimal localization. The reliance on stitched pipelines introduces information bottlenecks during backpropagation, which dilutes spatial signals and is limited by semantic priors of the segmentor. To address these limitations, we propose ForensicsTok, which reformulates image manipulation localization as an autoregressive sequence generation task. ForensicsTok directly generates spatially grounded token sequences, enabling precise mask prediction without intermediary supervision. Specifically, we introduce a Token Splatting Decoder (TSD) to map tokens to binary masks via codebook-aware code smoothing, which mitigates sharp gradients from deterministic detokenizers. Furthermore, to capture diverse tampering clues, we propose a Hierarchical Expert Fusion (HEF) module that injects multi-scale features from a forensic expert model. This unified architecture effectively compensates for the lack of forensic priors in standard MLLMs. Extensive experiments on six benchmarks show that ForensicsTok substantially improves over existing MLLM-based baselines and slightly improves over strong forensic expert baselines, while exhibiting stronger robustness to perturbations.\par\vspace{0.5em}
\end{abstract}

\section{Introduction}
\label{sec:intro}

\begin{figure}[t] 
\centering
\includegraphics[width=\linewidth]{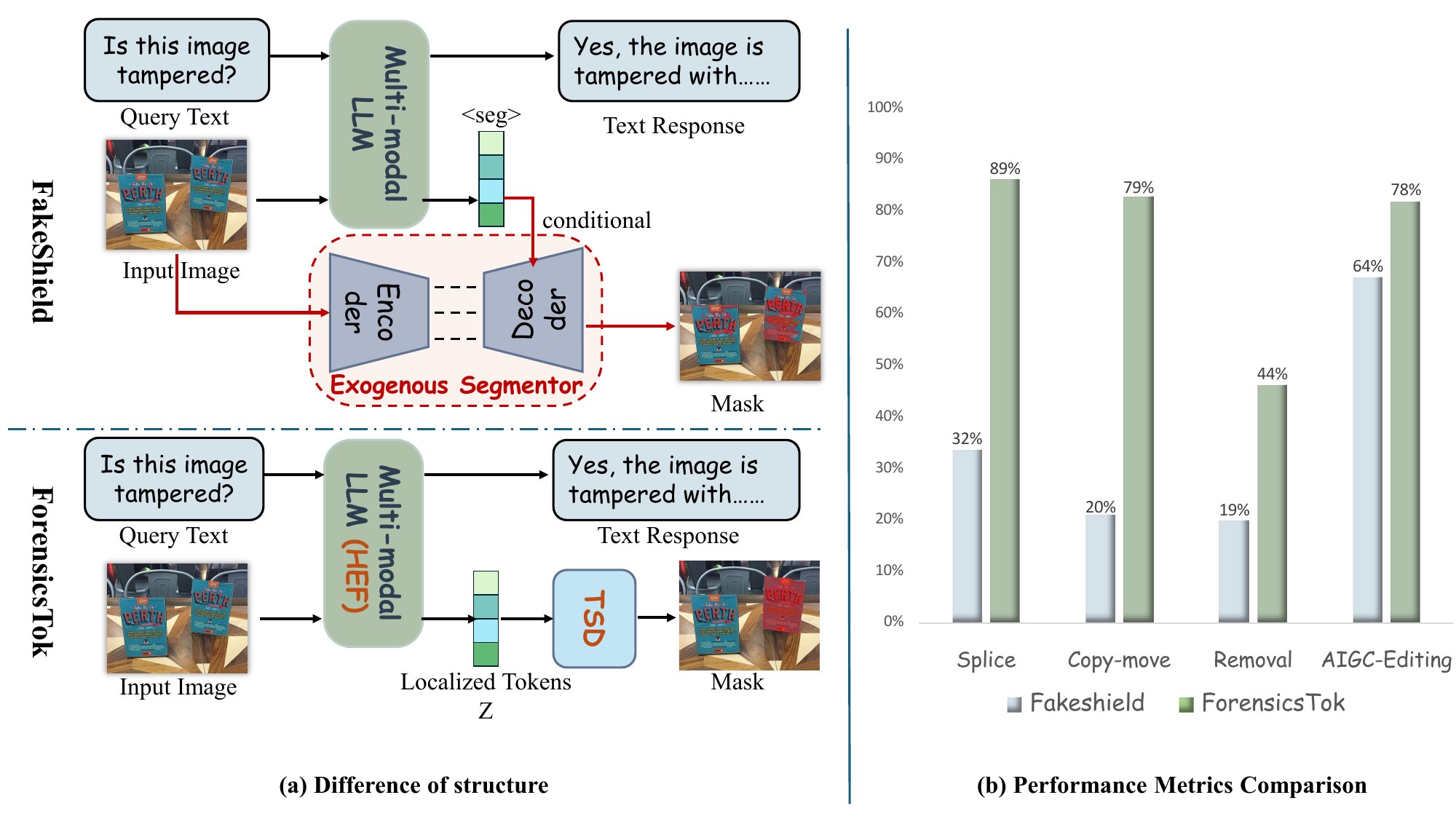} 
\caption{Comparison of FakeShield~\cite{xu2025fakeshield} and ForensicsTok. (a) FakeShield risks information loss via external decoders; ForensicsTok uses direct token generation for probabilistic masks. (b) We compared two methods under unified training protocols. F1 scores by tampering type, with ForensicsTok's average 39\% gain over FakeShield.}
\label{fig:structure_comparison}
\end{figure}

Beyond threats from traditional tampered images, the rapid advancement of AI-generated content (AIGC) has intensified risks like DeepFakes \cite{mirsky2021creation}, endangering public trust, legal proceedings, and information security.
Although prior expert models, such as ManTra-Net \cite{wu2019mantra} and SparseViT \cite{su2025can}, perform well on specific datasets like CASIA \cite{dong2013casia}, they struggle with highly diverse or semantically complex tampering scenarios, as shown by cross-dataset performance gaps.
Inspired by the powerful and consistent inference of multi-modal large language models (MLLMs), recent methods, such as FakeShield \cite{xu2025fakeshield} and SIDA \cite{huang2025sida}, employ them as reasoning detectors, framing IMDL (image manipulation detection and localization) as a prompt-based task.
To enable fine-grained localization, these methods integrate off-the-shelf MLLMs with segmentation decoders (\eg, SAM-like models \cite{kirillov2023segment}), raising two key concerns:

(1) \textbf{Ambiguous Representation of Tampering Mask Tokens}.
Existing methods generate tampering mask indirectly via external decoders with semantic embeddings from MLLM output.
These MLLM embeddings carry the semantic information of the masked region, disregarding the accurate spatial information.
The MLLM embeddings may refer to an ambiguous object (\eg, one among many of the same objects) in the images and cannot be precisely supervised by semantic labels.
Thus, adapting the existing framework in generating the tampering mask tokens is imprecise.

(2) \textbf{Inefficient Fusion of Forensic Knowledge}. 
Existing methods like FakeShield and SIDA rely solely on LoRA-based fine-tuning without integrating tampering-specific visual features, limiting their performance in locating fine-grain forensic traces.
Moreover, some recent approaches that incorporate forensic features through post-backbone multi-scale fusion strategies (\eg, AIGI-Holmes \cite{zhou2025aigi}) simply append them after the visual encoder.
However, the late-stage forensic features and visual features are heterogeneous.
Simple fusion of these features without multi-scale alignment is ineffective.

In this work, we propose an MLLM-based IML model by reformulating the IMDL task as an autoregressive sequence generation task and exploiting discriminative forensic features.
To allow precise representation of mask tokens, we introduce our ForensicsTok framework that directly generate spatially grounded token sequences from the MLLM.
It facilitates precise and supervised tampering mask prediction without ambiguous representation.
Specifically, inspired by TiTok~\cite{yu2024image}, we adopt a pre-trained codebook to map tokens to binary masks, ensuring explicit spatial modeling and end-to-end alignment through coarse to fine decoding.
Additionally, to effectively supervise the mask token, we introduce the Token Splatting Decoder (TSD) via codebook-aware label smoothing, stabilizing gradients by accounting for semantic similarities in the codebook.
To effectively integrate the IMDL-specific knowledge, we introduce the Hierarchical Expert Fusion (HEF) module to inject multi-scale forensic features from expert models into the visual backbone.
Specifically, this module aligns features at intermediate layers via attention-based fusion.
It preserves subtle cues while mitigating the heterogeneous feature issues of the post-backbone strategies.
As shown in \cref{fig:structure_comparison}(a), our ForensicsTok overcomes the existing limitations in mask token representation and IML feature fusion by eliminating external decoders. 
Consequently, as illustrated in \cref{fig:structure_comparison}(b), it achieves consistent localization performance with average F1 gains of 0.39 across various tampering types (from FakeShield's 0.33 to 0.72) under our unified protocol.
Extensive experiments further demonstrate that ForensicsTok achieves strong performance on multiple tampering localization datasets, with substantial improvements over existing MLLM-based models and modest but consistent gains over strong forensic experts, while exhibiting robustness to noise perturbations and cross-domain scenarios.
In summary, our main contributions are:

\begin{itemize}
\item We propose a tokenized autoregressive framework for IML, named ForensicsTok. Our Token Splatting Decoder (TSD) allows direct supervision and precise representation of tampering mask tokens.
\item We introduce the Hierarchical Expert Fusion (HEF) module to incorporate multi-scale tampering forensic features into the visual encoder in MLLM, enhancing perception of diverse tampering types.
\item We provide an empirical study showing that an MLLM-based localization pipeline can match or slightly surpass strong forensic expert models under challenging cross-domain protocols, while retaining a text-conditioned modeling interface.
\end{itemize}

\section{Related Works}

\subsection{Image Manipulation Detection \& Localization}
IMDL methods aim to generate precise tampering region masks.
For deep learning-based IMDL models, the performance improved significantly by data-driven techniques (\eg, ManTra-Net \cite{wu2019mantra}, MVSS-Net \cite{chen2021image}, and ObjectFormer \cite{wang2022objectformer}).
More recent expert models, such as SparseViT \cite{su2025can}, refine attention to suspicious regions and achieve strong IMDL performance.
Nevertheless, while most forensic methods are lightweight and fast, offering strong perception of low-level information such as texture inconsistencies, they lack broader world knowledge or semantic priors, restricting their ability to handle highly realistic or semantically complex tampering scenarios.
In this work, we integrate expert forensic knowledge into an MLLM-compatible localization pipeline; textual explanations are a potential use of this interface but are not the primary evidence evaluated in this paper.

\subsection{Multimodal Large Language Models}
To adapt existing LLMs for detection and grounding tasks, existing MLLM-based methods (\eg, Pix2Seq \cite{chen2022pix2seq}, SEEM \cite{zou2023seem}) align visual features and text prompts via cross-modal mechanisms to generate serialized outputs like masks.
These approaches typically extract image features with visual encoders, fuse them with text embeddings via projection layers, and feed the result into decoders for sequence prediction and mask reconstruction.
Recent efforts also empower MLLMs for forensic applications, such as explaining synthesized images \cite{he2025can}.
To allow reasoning-driven segmentation based on the input instruction, LISA \cite{lai2024lisa} uses a \verb|<seg>| token to encode hidden embeddings into masks and facilitate zero-shot generalization for different queries.
To boost the performance in complex scenes, HiMTok \cite{wang2025himtok} tokenizes visual signals into 1D sequences for MLLM-predicted multi-layer masks.
For MLLM-based IMDL methods, initial attempts such as FakeShield \cite{xu2025fakeshield} and SIDA \cite{huang2025sida} follow the LISA framework by coupling MLLMs with external decoders (\eg, SAM \cite{kirillov2023segment}) to generate masks from hidden embeddings.
However, as highlighted in \cref{sec:intro}, these approaches \cite{xu2025fakeshield, huang2025sida} suffer from the limitations of ambiguous representation of mask tokens and inefficient fusion of forensic knowledge.
To overcome these limitations in the forensic setting, we build ForensicsTok, which directly predicts discrete token sequences through MLLMs for mask generation with direct supervision of the Token Splatting Decoder (TSD), while incorporating expert forensic features with the Hierarchical Expert Fusion (HEF) module. We emphasize that the codebook-based mask tokenization itself follows prior tokenized segmentation work; the forensic contributions lie in the IMDL adaptation, codebook-aware smoothing, and intermediate expert fusion.

\section{Methodology}

\begin{figure}[t]
\centering
\includegraphics[width=\linewidth]{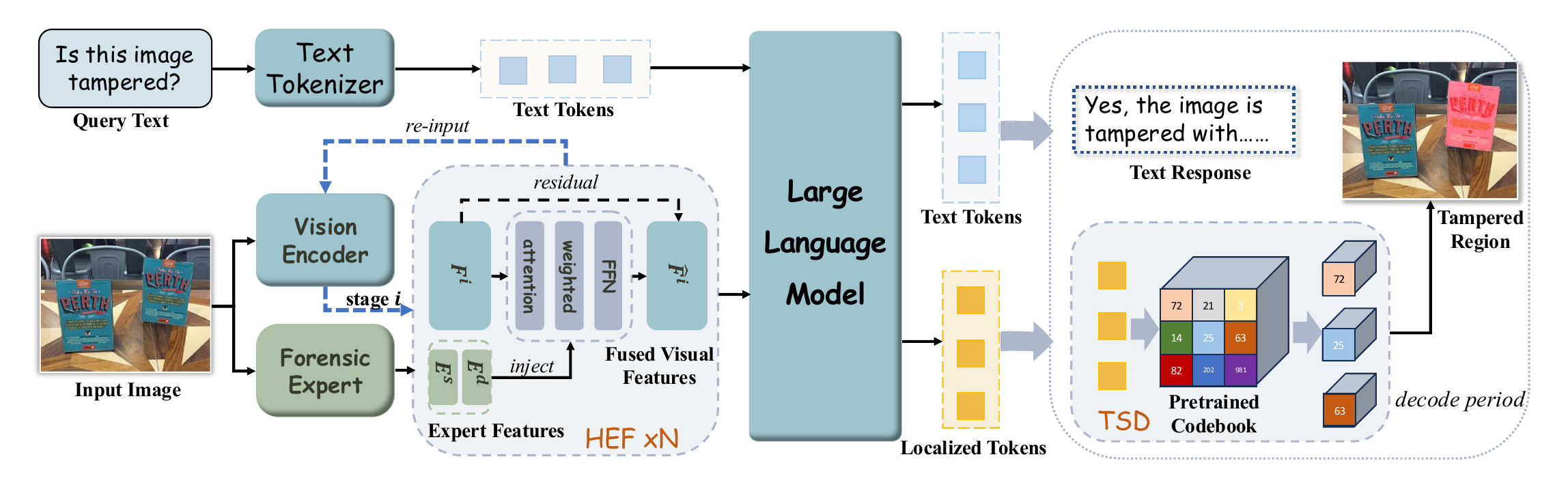}
\caption{Overview of the proposed ForensicsTok framework. The architecture integrates the Hierarchical Expert Fusion (HEF) module with the visual backbone, which efficiently incorporates expert forensic knowledge for tampering localization.
ForensicsTok also employs the Token Splatting Decoder (TSD), a forensic-specific token supervision and detokenization path that maps tokens to binary masks.}
\label{fig:overview}
\end{figure}

\subsection{ForensicsTok}
Existing MLLM-based methods~\cite{xu2025fakeshield, huang2025sida} utilize off-the-shelf segmentor~\cite{kirillov2023segment} to locate tampered regions with the semantic outputs from MLLM.
The stitched design brings architectural overhead and prevents direct supervision from spatial annotations, suffering from information bottlenecks during backpropagation.
To address those problems, we build ForensicsTok, which reformulates localization as detokenizing an autoregressive sequence as shown in \cref{fig:overview}.
To enhance its visual encoder with prior knowledge of tampering, we first introduce the Hierarchical Expert Fusion (HEF).
The purpose of HEF is to take the various advantages of forgery experts according to different tampering methods.
Specifically, we propose a multi-scale gate mechanism to fuse the parallel expert features and original visual features adaptively.
For localization, we adopt a codebook-based mask detokenization paradigm and introduce a forensic-specific supervision path, Token Splatting Decoder (TSD).
Concretely, the hard codebook detokenizer is inherited from tokenized segmentation methods~\cite{wang2025himtok, yu2024image}; it maps generated mask tokens to binary tampering probabilities.
Our added objective is codebook-aware label smoothing ($L_{\text{TSD}}$), which redistributes a small probability mass from the ground-truth code to its semantic neighbors in the fixed codebook.
This explicitly addresses the rigid one-hot supervision issue of binary-mask codebooks and stabilizes optimization for IMDL.

\subsection{Token Splatting Decoder}
\label{sec:TSD}

\begin{figure}[tb]
\centering
\includegraphics[width=\linewidth]{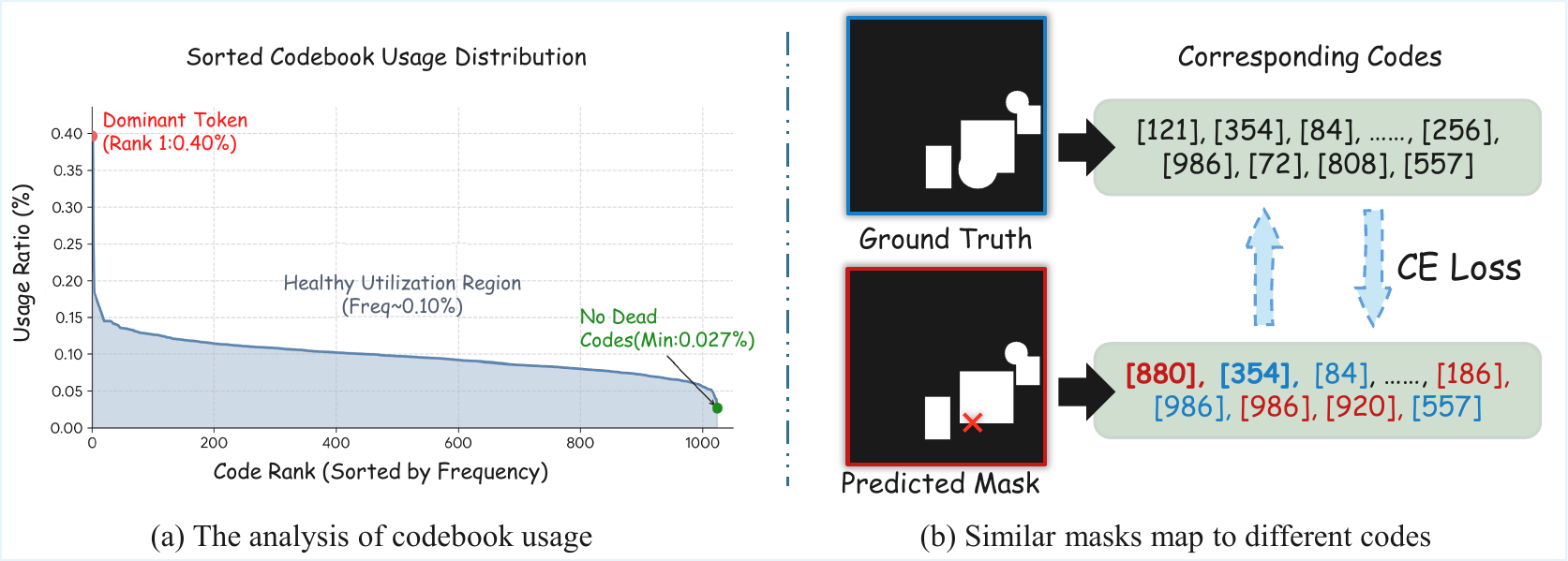}
\caption{Codebook usage and rigid supervision. 
(a) Code usage is balanced. 
(b) Distinct sequences can decode to similar masks. Treating such semantically-close codes as exclusive one-hot labels causes strict supervision and instability.}
\label{fig:TSD}
\end{figure}

The original MLLM architecture is suitable for autoregressive token generation but is limited when localizing 2D masks. 
Existing MLLM-based IMDL methods often attach external segmentors to sparse semantic tokens, making mask prediction depend on hidden semantic embeddings rather than directly supervised mask tokens. We instead represent tampering masks as discrete tokens for autoregressive generation. Since visually similar local masks may correspond to neighboring codebook indices, standard one-hot supervision can be overly rigid. TSD addresses this mismatch by smoothing supervision over semantically neighboring codebook entries while keeping hard decoding unchanged at inference.

\paragraph{Codebook-based Detokenizer.}
HiMTok~\cite{wang2025himtok} utilizes a codebook-based detokenizer to map discrete tokens to binary masks. We use this mechanism as the tokenized mask interface rather than claiming it as a new tokenization design.
Formally, given a token feature $\bm{v}$ and a pre-trained codebook $\mathcal{C}=\{e_k\}_{k=1}^{N}$, the inference process performs hard assignment via:
\begin{equation}
\hat{y} = \argmin_k \| \bm{v} - e_k \|_2,
\label{eq:codebook}
\end{equation}
where $\hat{y}$ is the selected code index. A sequence of $\hat{y}$ is then decoded into the final mask.

\paragraph{Rigid Token Supervision.}
While \cref{eq:codebook} is not involved in backpropagation, standard one-hot supervision over discrete codes introduces optimization instability.
As illustrated in \cref{fig:TSD}(b), visually similar masks often map to adjacent codes in the embedding space.
However, the standard cross-entropy objective treats all non-target codes as orthogonal. 
For two semantically codes $y$ and $y'$, the loss difference can be disproportionately large:
\begin{equation}
\Delta \mathcal{L}_{\text{CE}}
= \mathcal{L}_{\text{CE}}(\bm{z}, y) - \mathcal{L}_{\text{CE}}(\bm{z}, y')
= -\log p_y + \log p_{y'},
\end{equation}
where $\bm{z}$ denotes the model logits and $p=\text{Softmax}(\bm{z})$.
This ``all-or-nothing'' penalty ignores the geometric structure of the codebook, leading to unstable gradient signals.

\paragraph{Codebook-aware Label Smoothing ($L_{\text{TSD}}$).}
We instantiate this idea as a target-driven label smoothing mechanism on top of the fixed codebook detokenizer.
For a ground-truth code $y \in \mathcal{C}$, where each code $j \in \mathcal{C}$ is associated with an embedding $e_j \in \mathbb{R}^d$,
we compute similarity scores $s_j = \text{Sim}(e_y, e_j)$ for all $j \in \mathcal{C}$ (with $\text{Sim}(\cdot,\cdot)$ implemented as cosine similarity).
Let $\pi_y$ be a permutation of $\mathcal{C}$ satisfying $\pi_y(1)=y$, and
\begin{equation}
s_{\pi_y(2)} \ge s_{\pi_y(3)} \ge \cdots \ge s_{\pi_y(|\mathcal{C}|)},
\end{equation}
where $s_j=\text{Sim}(e_y,e_j)$ for all $j\in\mathcal{C}$.
Then we define the semantic neighborhood as
\begin{equation}
\mathcal{N}_{K}(y) = \{\pi_y(2), \pi_y(3), \ldots, \pi_y(K)\}.
\end{equation}
Consequently, the effective neighborhood size is $|\mathcal{N}_{K}(y)| = K-1$. So choose any $k$ from $\mathcal{N}_K(y)$, we assign smoothing weights to these neighbors using a temperature-scaled softmax:
\begin{equation}
w_{k} =
\frac{\exp(\text{Sim}(e_y, e_k)/\tau)}
{\sum_{j\in \mathcal{N}_{K}(y)} \exp(\text{Sim}(e_y, e_j)/\tau)}.
\end{equation}
The final TSD loss is formulated as a mixture of the hard target loss and the weighted neighbor loss:
\begin{equation}
L_{\text{TSD}}(\bm{z}, y) =
(1-\epsilon)\,\mathcal{L}_{\text{CE}}(\bm{z}, y)
+ \epsilon \sum_{k\in \mathcal{N}_{K}(y)} w_{k}\, \mathcal{L}_{\text{CE}}(\bm{z}, k),
\label{eq:TSD}
\end{equation}
where $\epsilon$ is a smoothing factor.
In the case of $K=1$, the set $\mathcal{N}_{K}(y)$ is empty, reducing \cref{eq:TSD} to standard cross-entropy.

Since the neighborhood $\mathcal{N}_{K}(y)$ and weights $w_{k}$ are derived solely from the fixed codebook and ground truth, they are constant with respect to $\bm{z}$, which provides naturally smoother supervision signals without differentiating through the similarity kernel.

\subsection{Hierarchical Expert Fusion}

\begin{figure}[t] 
  \centering
  \includegraphics[width=1.0\columnwidth]{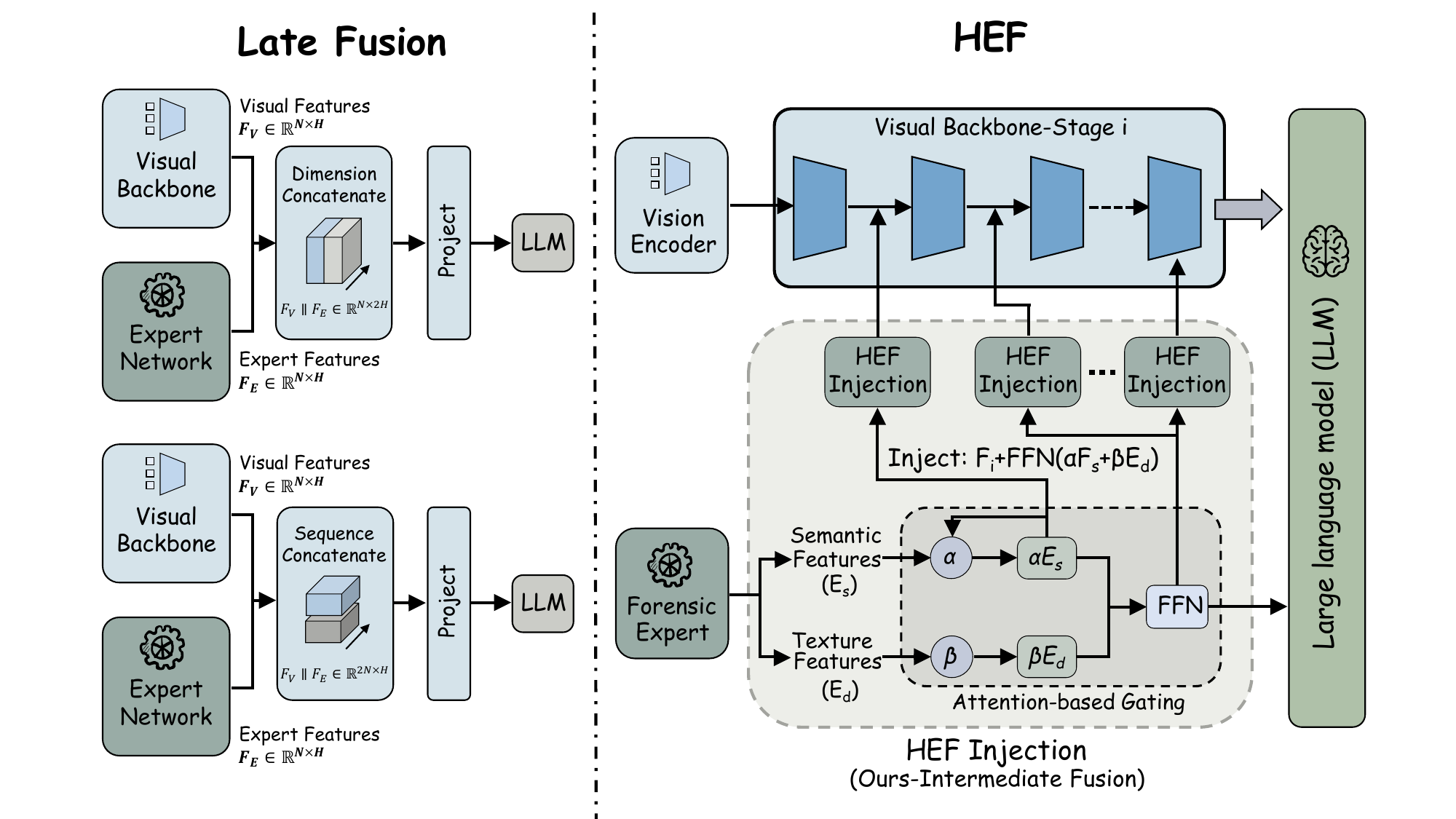} 
  
  \caption{\textbf{Comparison of Feature Fusion Architectures.} 
  \textbf{Left (Late Fusion):} Existing methods (\eg, Dimension or Sequence Concatenation) simply fuse forensic features after the visual backbone.
  \textbf{Right (Ours):} The proposed \textbf{Hierarchical Expert Fusion (HEF)} adopts an intermediate fusion strategy.}
  \label{fig:HEF_architecture} 
\end{figure}

Although an MLLM offers semantic and text-conditioned modeling, its visual backbone is not trained to preserve the low-level artifacts that often determine image manipulation localization.
Conversely, forensic experts capture compression traces, boundary discontinuities, and texture inconsistencies, but they provide limited semantic reasoning and are sensitive to the training distribution.
HEF is therefore designed as an intermediate fusion interface: it injects multi-scale expert features into the visual encoder before the LLM projection, rather than concatenating heterogeneous features only after the visual backbone.
This design is motivated by the fact that different manipulations require different cues, such as low-level texture for removal or boundary artifacts, and more semantic inconsistency for copy-move or object-level splicing.

We use SparseViT~\cite{su2025can} as the frozen expert because it is a strong recent forensic model and has a feature hierarchy compatible with the visual encoder.
The role of HEF is not to make the framework independent of expert quality; weak or poorly aligned experts can inject noisy cues, as analyzed in \cref{tab:ablation_expert_selection}.
Instead, HEF provides a gated mechanism to exploit a reliable expert prior when available.
With an image $\bm{I}$, we first use the expert to extract two progressive features $\bm{E}_s$ and $\bm{E}_d$, where the former is from a shallow layer and the latter is from a deeper layer.
For the $i$-th stage, the features from visual backbone are formulated as $\bm{F}^i$.
We first align the spatial and channel dimensions of the expert features with the visual-backbone features:

\begin{equation}  
\tilde{\bm{E}}_s = \text{resize}(\bm{E}_s, \, \text{size}(\bm{F}^i)), \quad
\tilde{\bm{E}}_d = \text{resize}(\bm{E}_d, \, \text{size}(\bm{F}^i)),
\end{equation}
where $\text{size}(\cdot)$ denotes the target feature shape.
After that, we compute the attention weight of two expert features as:
\begin{equation}  
[\alpha^i, \beta^i] = \text{Softmax}\left(\frac{(W_q^i \bm{F}^i)^\top \, [W_s^i\tilde{\bm{E}}_s,W_d^i\tilde{\bm{E}}_d]}{\sqrt{d}}\right),
\end{equation}
where $\alpha^i$ and $\beta^i$ denote normalized weights for shallow and deep expert information, respectively, and $W$ denotes learnable projection matrices.
Then we integrate all the expert information with the weight to form the sparse-enhanced feature, which can be described as:

\begin{equation}    
\hat{\bm{F}^i} = \bm{F}^i+\text{FFN}(\alpha^i \tilde{\bm{E}}_s + \beta^i \tilde{\bm{E}}_d),
\end{equation}
where $\hat{\bm{F}^i}$ continues to forward pass through the original encoder structure and is ultimately being fed into the LLM with the textual features.
Through the HEF process outlined above, we are able to selectively inject multi-scale expert information at each stage of the visual encoder.
HEF further deconstructs the expert internally, enabling the new feature $\hat{\bm{F}^i}$ to more flexibly address various types of tampering.
For instance, in the case of mosaic tampering, which disrupts image pixels, HEF places greater emphasis on capturing low-level textures.
In contrast, for copy-move tampering, which has fewer traces of manipulation, HEF emphasizes the extraction of disharmonious semantics.
The final loss combines token supervision, mask reconstruction, and hierarchical mask constraints, as described below.

\subsection{Training Objectives}
\label{sec:training_objectives}
We train ForensicsTok with a unified objective that jointly supervises token prediction and mask reconstruction.
The overall training objective is defined as a linear combination of three losses:
\begin{equation}
\label{eq:total_loss}
L_{\text{total}}
=
\lambda_{\text{TSD}}\,L_{\text{TSD}}
+
\lambda_{\text{mask}}\,L_{\text{mask}}
+
\lambda_{\text{HML}}\,L_{\text{HML}},
\end{equation}
where we set $\lambda_{\text{TSD}}=\lambda_{\text{mask}}=\lambda_{\text{HML}}=1$ in all experiments.

\noindent\textbf{Token Splatting Decoder Loss.}
$L_{\text{TSD}}$ is the codebook-aware label smoothing objective used to supervise discrete mask tokens,
which redistributes a small probability mass from the ground-truth code to its semantic neighbors in the fixed codebook,
stabilizing optimization. The formulation follows \cref{eq:TSD}.

\noindent\textbf{Mask Loss.}
Let $\hat{M}$ and $M$ denote the predicted and ground-truth tampering masks, respectively.
We apply a pixel-level supervision composed of binary cross-entropy and Dice loss:
\begin{equation}
\label{eq:mask_loss}
L_{\text{mask}}
=
L_{\text{BCE}}(\hat{M}, M)
+
\gamma\,L_{\text{Dice}}(\hat{M}, M)
\end{equation}
where the Dice coefficient weight $\gamma$ is fixed at 0.25 throughout training.

\noindent\textbf{Hierarchical Mask Loss.}
We further adopt the hierarchical mask loss inherited from HiMTok~\cite{wang2025himtok},
which complements the above objectives by imposing hierarchical constraints on mask prediction.

\section{Experiments}

\subsection{Experimental Setup}
\label{subsec:Experimental Setup}

\noindent\textbf{Dataset}
To ensure a fair and comprehensive evaluation, we design an experimental protocol that emphasizes cross-domain generalization for MLLMs, differing from protocols tailored for specialized forensic models (\eg, SparseViT~\cite{su2025can}), which often neglect AIGC-based manipulations, and from in-domain evaluations in SIDA~\cite{huang2025sida}.
Our protocol incorporates a broader mix of datasets to enhance coverage and robustness across diverse tampering scenarios.

Our training process utilizes TamperCOCO~\cite{kwon2021cat}, MIML~\cite{qu2024towards}, CASIA2~\cite{dong2013casia}, and tampered samples from SID\_Set~\cite{huang2025sida}.
These datasets---TamperCOCO, MIML, and CASIA2---are commonly used in IMDL research~\cite{ma2024imdl, qu2024towards, nandi2023trainfors}.
Additionally, SID\_Set, focused on AIGC tampering, strengthens the model's exposure to AI-generated forgeries.
Evaluation is conducted on six diverse benchmarks to assess cross-domain localization and generalization on unseen datasets: CASIA1~\cite{dong2013casia}, NIST~\cite{yates2017nimble}, Coverage~\cite{wen2016coverage}, Columbia~\cite{ng2009columbia}, Glide~\cite{guillaro2023trufor}, and IMD~\cite{novozamsky2020imd2020}.
These encompass various tampering operations (\eg, splicing, copy-move, removal, and AIGC-based forgeries).
None of the six evaluation benchmarks is used for training or validation in our experiments.
For CASIA, we use the released CASIA2 subset only for training and CASIA1 only for testing; no CASIA1 image or mask is used during training.
Because CASIA1 and CASIA2 share the same dataset family, we interpret CASIA1 as a disjoint within-family test and rely on NIST, Coverage, Columbia, Glide, and IMD for stronger cross-source evidence.

\noindent\textbf{Implementation Details}
We initialize the model with pre-trained weights from InternVL2.5-8B~\cite{chen2024expanding} for the MLLM backbone and HiMTok~\cite{wang2025himtok} for the decoder/codebook, while adopting SparseViT~\cite{su2025can} as a fixed expert feature extractor.
SparseViT is used only to provide intermediate features; its predicted masks are not used as pseudo-labels.
During training, we fine-tune the MLLM backbone together with our introduced modules (\eg, HEF and the decoder parameters), while keeping SparseViT frozen.
We use AdamW optimizer with a base learning rate of 2e-5 (increased to 4e-5 for the Hierarchical Expert Fusion (HEF) module to accelerate adaptation), weight decay of 0.05. Images are resized to 448$\times$448 resolution with random flipping and color jittering for augmentation, alongside dynamic image sizing, thumbnail usage, and data resampling.
Unless otherwise noted, localization baselines in \cref{tab:loc} are trained or fine-tuned under the same training pool and evaluated using the same preprocessing and metrics.
Consequently, the numbers for methods such as FakeShield may differ from those in their original papers, which used different data splits, preprocessing, thresholds, or evaluation protocols; \cref{tab:loc} should be read as a controlled comparison within our protocol rather than a reproduction table for each baseline.

\noindent\textbf{Evaluation Metrics}
Following prior IML literature, we emphasize pixel-level metrics: Intersection over Union (IoU) and F1 score, which measure overlap and accuracy between predicted and ground-truth tampering masks.

\subsection{Localization Performance Results}

\begin{table}[t] 
\centering
\caption{Comparison of tampering localization performance for ForensicsTok against traditional forensic methods and recent MLLM-based approaches, evaluated using IoU and F1 metrics across six datasets.
Bold and underlined values indicate the best and second-best performance, respectively, among all competing methods.}
\label{tab:loc}
\resizebox{\textwidth}{!}{%
\begin{tabular}{l|cccccccccccc|cc}
\toprule
\multirow{2}{*}{\centering Method} & \multicolumn{2}{c}{CASIA1} & \multicolumn{2}{c}{NIST} & \multicolumn{2}{c}{Coverage} & \multicolumn{2}{c}{Columbia} & \multicolumn{2}{c}{Glide} & \multicolumn{2}{c|}{IMD} & \multicolumn{2}{c}{Average} \\
\cmidrule(lr){2-3} \cmidrule(lr){4-5} \cmidrule(lr){6-7} \cmidrule(lr){8-9} \cmidrule(lr){10-11} \cmidrule(lr){12-13} \cmidrule(lr){14-15}
 & IoU & F1 & IoU & F1 & IoU & F1 & IoU & F1 & IoU & F1 & IoU & F1 & IoU & F1 \\ 
\midrule
\footnotesize{\textit{\textbf{Forensic Expert}}} &  &  &  &  &  & & & & & & & & & \\ 
CAT-Net~\cite{kwon2021cat}    & 0.38 & 0.56 & 0.29 & 0.33 & 0.13 & 0.23 & 0.77 & 0.77 & 0.28 & 0.43 & 0.30 & 0.34 & 0.36 & 0.44 \\ 
MVSS-Net~\cite{chen2021image}   & 0.49 & 0.70 & 0.25 & 0.30 & 0.43 & 0.59 & 0.86 & \underline{0.90} & 0.44 & 0.62 & 0.40 & 0.53 & 0.48 & 0.61 \\ 
PSCC-Net~\cite{liu2021pscc}   & 0.29 & 0.43 & 0.20 & 0.26 & 0.30 & 0.43 & 0.38 & 0.54 & 0.39 & 0.58 & 0.17 & 0.27 & 0.29 & 0.42 \\ 
TruFor~\cite{guillaro2023trufor}     & 0.60 & 0.74 & 0.36 & 0.43 & 0.46 & 0.50 & 0.75 & 0.78 & 0.48 & 0.66 & 0.48 & 0.59 & 0.52 & 0.62 \\ 
SparseViT~\cite{su2025can}  & \underline{0.68} & \underline{0.82} & \underline{0.39} & \underline{0.57} & \underline{0.62} & \textbf{0.79} & \underline{0.89} & 0.88 & \underline{0.65} & \underline{0.77} & \underline{0.60} & \underline{0.73} & \underline{0.64} & \underline{0.76} \\ 
\midrule
\footnotesize{\textit{\textbf{Forensic MLLM}}} &  &  &  &  &  & & & & & & & & & \\ 
FakeShield~\cite{xu2025fakeshield} & 0.48 & 0.68 & 0.17 & 0.22 & 0.27 & 0.46 & 0.68 & 0.75 & 0.47 & 0.64 & 0.02 & 0.03 & 0.35 & 0.46 \\ 
SIDA~\cite{huang2025sida}       & 0.46 & 0.63 & 0.27 & 0.43 & 0.30 & 0.45 & 0.65 & 0.70 & 0.48 & 0.58 & 0.46 & 0.61 & 0.44 & 0.57 \\ 
\rowcolor{yellow!15}
ForensicsTok & \textbf{0.78} & \textbf{0.85} & \textbf{0.44} & \textbf{0.58} & \textbf{0.64} & \underline{0.77} & \textbf{0.89} & \textbf{0.90} & \textbf{0.66} & \textbf{0.80} & \textbf{0.65} & \textbf{0.76} & \textbf{0.68} & \textbf{0.78} \\ 
\bottomrule
\end{tabular}%
}
\end{table}

\cref{tab:loc} presents a comprehensive comparison on the tampering localization performance of our ForensicsTok against both forensic expert models (\eg, CAT-Net \cite{kwon2021cat}, MVSS-Net \cite{chen2021image}, PSCC-Net\cite{liu2021pscc}, TruFor\cite{guillaro2023trufor}, SparseViT \cite{su2025can}) and MLLM-based methods (\eg, FakeShield \cite{xu2025fakeshield}, SIDA \cite{huang2025sida}).
All compared methods were trained on the same datasets outlined in \cref{subsec:Experimental Setup}.
ForensicsTok achieves the highest average IoU (0.68) and F1 (0.78) scores in this protocol.
Compared to forensic expert models, ForensicsTok improves over the top baseline SparseViT (average IoU 0.64, F1 0.76) by 0.04 in IoU and 0.02 in F1.
We view this gain as modest rather than conclusive evidence that MLLMs replace forensic experts; importantly, ForensicsTok also uses SparseViT as a frozen feature prior.
The main empirical advantage is instead that direct mask-token generation closes much of the gap between MLLM-based localization and specialized expert models.
The gains are particularly pronounced on CASIA1 (+0.10 IoU, +0.03 F1), where splicing forgeries often contain semantically distinct manipulations.
In contrast, our gains on Coverage are marginal because this dataset emphasizes copy-move forgeries, where non-semantic fine-grained traces such as texture inconsistencies are critical and forensic experts already perform strongly.
Against MLLM-based methods, ForensicsTok significantly outpaces SIDA (average IoU 0.44, F1 0.57) by 0.24 in IoU and 0.21 in F1, and FakeShield (0.35/0.46) by 0.33/0.32.
These improvements mainly arise from reformulating IMDL as an autoregressive token generation problem, which allows direct supervision of MLLM output tokens and avoids the information bottleneck and semantic mismatch introduced by exogenous segmentation decoders.

\subsection{Detection performance results}

We adhere to the experimental settings outlined in the SIDA paper~\cite{huang2025sida} for evaluations on the SID-Set dataset.
The dataset is partitioned into training, validation, and test sets in a 7:1:2 ratio.
All models presented in the table are trained on the SID-Set training split and evaluated on the test set.
In line with the detection metrics employed in~\cite{huang2025sida}, we report per-class Accuracy (ACC) and F1 scores, along with their macro-averaged values across the three classes, denoted as ``Overall.''  For classification, the SIDA model directly outputs category predictions, whereas all other models derive predictions from their output masks: a fully background mask indicates ``Real,'' a fully foreground mask indicates ``Fully Synthetic,'' and any intermediate mask indicates ``Tampered.''

\begin{table}[t] 
\centering
\caption{Classification performance on \textbf{SID-Set} using the SIDA protocol.
We report per-class Accuracy (ACC) and F1, and their macro averages (Overall); the macro-averaged ACC is equivalent to a class-balanced accuracy over the three classes.}
\label{tab:cls_sidset}
\begin{tabular}{c|cccccc|cc}
\toprule
\multirow{2}{*}{\centering Method} & \multicolumn{2}{c}{Real} & \multicolumn{2}{c}{Fully Synthetic} & \multicolumn{2}{c|}{Tampered} & \multicolumn{2}{c}{Overall} \\
\cmidrule(lr){2-3} \cmidrule(lr){4-5} \cmidrule(lr){6-7} \cmidrule(lr){8-9}
& ACC & F1 & ACC & F1 & ACC & F1 & ACC & F1 \\
\midrule
\footnotesize{\textit{\textbf{Forensic Expert}}} & & & & & & & & \\
MVSS-Net~\cite{chen2021image}      & 0.98 & 0.98 & 0.97 & 0.96 & 0.97 & 0.95 & \textbf{0.97} & 0.97 \\
PSCC-Net~\cite{liu2021pscc}      & \textbf{0.99} & \textbf{0.99} & 0.36 & 0.53 & 0.01 & 0.03 & 0.45 & 0.52 \\
TruFor~\cite{guillaro2023trufor}    & 0.73 & 0.84 & \textbf{0.99} & \textbf{0.99} & \textbf{0.98} & \textbf{0.99} & 0.90 & 0.94 \\
SparseViT~\cite{su2025can} & 0.96 & 0.98 & \textbf{0.99} & \textbf{0.99} & 0.93 & 0.96 & 0.96 & \textbf{0.98} \\
\midrule
\footnotesize{\textit{\textbf{Forensic MLLM}}} & & & & & & & & \\
SIDA~\cite{huang2025sida}      & 0.89 & 0.91 & 0.98 & 0.98 & 0.92 & 0.91 & 0.93 & 0.93 \\
\rowcolor{yellow!15}
ForensicsTok & \textbf{0.99} & 0.97 & \textbf{0.99} & \textbf{0.99} & 0.94 & 0.97 & \textbf{0.97} & \textbf{0.98} \\
\bottomrule
\end{tabular}
\end{table}

As shown in \cref{tab:cls_sidset}, ForensicsTok attains strong overall performance, with macro-averaged ACC and F1 scores of 0.97 and 0.98, respectively.
It surpasses SIDA (0.93/0.93) and is comparable to SparseViT (0.96/0.98), but the gain over the strongest expert is limited.
This result supports the usefulness of the tokenized MLLM formulation for unified detection/localization, while also showing that lightweight forensic experts remain highly competitive for pure classification.
Relative to SIDA, ForensicsTok exhibits gains in all classes, including a 0.10/0.06 improvement in ACC/F1 for the ``Real'' class and a 0.02/0.06 enhancement for the ``Tampered'' class.

\subsection{Ablation Study}

\subsubsection{Architecture Ablation}
As shown in \cref{tab:arch_ablation_iou}, we conduct an ablation study on the key components in our model.
Note that the baseline without any of the three components (\eg, the first row in the table) corresponds to the vanilla LISA design \cite{lai2024lisa}.

\textbf{Codebook-based Decoder Ablation}: 
Our variant with only the codebook-based decoder (No.2) improves performance to an average \textbf{IoU} of \textbf{0.61}, featuring a \textbf{0.32} gain over the LISA baseline (No.1).
It highlights that a HiMTok-style codebook decoder is a strong foundation for mask-token prediction in IMDL, and it accounts for the largest single gain in this ablation.

\textbf{Expert Fusion Ablation}: 
The variant with HEF (No.3) shows a performance gain of \textbf{0.03} in average \textbf{IoU} score over that of the second variant.
Coverage benefits the most (\textbf{+0.10}) among all datasets, emphasizing the role of HEF in detecting subtle copy-move inconsistencies.
Adding HEF reaches \textbf{0.64}, demonstrating gains from forensic feature extraction.

\textbf{Token Splatting Decoder Ablation}: 
The variant with token splatting (No.4) demonstrates a \textbf{0.04} gain in \textbf{IoU} score over that of the third variant.
The full model achieves \textbf{0.68}.
It confirms the benefits of codebook-aware smoothing for stabilizing token gradients.

\begin{table}[t]
\centering
\caption{Ablation study on component effectiveness. We report \textbf{IoU} across six datasets. (CBD: Codebook-based Decoder, HEF: Hierarchical Expert Fusion, TSD: Token Splatting Decoder).}
\label{tab:arch_ablation_iou}
\begin{tabular}{ccc|cccccc|c}
\toprule
CBD & HEF & TSD & 
CASIA1 & NIST & Coverage & Columbia & Glide & IMD & \textbf{Average} \\
\midrule 
\ding{55} & \ding{55} & \ding{55} & 
0.30 & 0.29 & 0.19 & 0.39 & 0.28 & 0.30 & 0.29 \\
\checkmark & \ding{55} & \ding{55} & 
0.73 & 0.40 & 0.47 & 0.86 & 0.61 & 0.60 & 0.61 \\
\checkmark & \checkmark & \ding{55} & 
0.75 & 0.38 & 0.57 & 0.88 & 0.65 & 0.62 & 0.64 \\
\rowcolor{yellow!15} 
\checkmark & \checkmark & \checkmark & 
\textbf{0.78} & \textbf{0.44} & \textbf{0.64} & \textbf{0.89} & \textbf{0.66} & \textbf{0.65} & \textbf{0.68} \\
\bottomrule
\end{tabular}%
\end{table}

\subsubsection{Feature Fusion Ablation}

To demonstrate the superiority of our intermediate fusion strategy in the HEF, we compare it against post-backbone fusion strategies (visualized in the Left panel of \cref{fig:HEF_architecture}), with the SparseViT branch fixed.
For sequence concatenation, we adopt the approach from \cite{zhou2025aigi}, which concatenates visual features token-wise, resulting in 2$N$ tokens fed to the LLM after projection.
For dimension concatenation, we implement a baseline variant that appends sparse features channel-wise followed by a 2$H$ to $H$ projection before feeding to the LLM.
As shown in \cref{tab:fusion_ablation}, our intermediate fusion achieves an average IoU of 0.68, outperforming the performances of other fusion strategies.
It is because both sequence and dimension concatenation are late-stage strategies, which dilute forensic cues through suboptimal integration between heterogeneous features.
In contrast, our intermediate fusion preserves multi-scale alignment, yielding notable gains on CASIA1 (+0.06 IoU vs. dimension concatenation) and Coverage (+0.12 IoU vs. dimension concatenation).

\begin{table}[t]
\centering
\caption{Ablation study on feature fusion strategies. We report \textbf{IoU} scores.}
\label{tab:fusion_ablation}
\begin{tabular}{l|cccccc|c}
\toprule
Method & CASIA1 & NIST & Coverage & Columbia & Glide & IMD & \textbf{Average} \\
\midrule
Dimension Concat & 0.72 & 0.33 & 0.52 & 0.89 & 0.59 & 0.61 & 0.61 \\
Sequence Concat  & 0.69 & 0.34 & 0.57 & 0.88 & 0.64 & 0.59 & 0.62 \\ 
\rowcolor{yellow!15}
HEF (Ours)       & \textbf{0.78} & \textbf{0.44} & \textbf{0.64} & \textbf{0.89} & \textbf{0.66} & \textbf{0.65} & \textbf{0.68} \\
\bottomrule
\end{tabular}
\end{table}

\subsubsection{Expert Model Selection Ablation}
We further investigated the rationale for selecting SparseViT as the forensic expert by replacing it with 
other expert models (MVSS-Net and PSCC-Net) within the HEF module.
The results in \cref{tab:ablation_expert_selection} demonstrate that SparseViT achieves the best absolute performance (0.68 Avg. IoU).
ForensicsTok+MVSS and ForensicsTok+PSCC improve substantially over their corresponding standalone experts, but they do not exceed the CBD-only variant in \cref{tab:arch_ablation_iou}.
This indicates that HEF is not independent of expert quality: a weaker or less aligned expert can introduce noisy cues that the gating mechanism cannot fully suppress.
We therefore interpret HEF as a mechanism for exploiting a reliable forensic prior rather than as an expert-agnostic performance guarantee.
Given that the differences in computational burden among these lightweight expert models are negligible relative to the MLLM backbone, we adopt SparseViT based on its superior feature quality.

\begin{table}[t]
    \centering
    \caption{Ablation study on expert model selection. We report \textbf{IoU} scores. We compare standalone experts vs. their integration into ForensicsTok.}
    \label{tab:ablation_expert_selection}
    \begin{tabular}{l|cccccc|c}
        \toprule
        Method & CASIA1 & NIST & Coverage & Columbia & Glide & IMD & \textbf{Average} \\
        \midrule
        \textit{Expert: MVSS} & 0.49 & 0.25 & 0.43 & 0.86 & 0.44 & 0.40 & 0.48 \\
        \textbf{ForensicsTok + MVSS} & 0.70 & 0.31 & 0.39 & \textbf{0.89} & 0.56 & 0.58 & 0.57 \\
        \midrule
        \textit{Expert: PSCC} & 0.29 & 0.20 & 0.30 & 0.38 & 0.39 & 0.17 & 0.29 \\
        \textbf{ForensicsTok + PSCC} & 0.73 & 0.39 & 0.45 & 0.86 & 0.59 & 0.59 & 0.60 \\
        \midrule
        \textit{Expert: SparseViT} & 0.68 & 0.39 & \textbf{0.62} & \textbf{0.89} & 0.65 & 0.60 & 0.64 \\
        \rowcolor{yellow!15}
        \textbf{ForensicsTok + SparseViT} & \textbf{0.78} & \textbf{0.44} & 0.64 & \textbf{0.89} & \textbf{0.66} & \textbf{0.65} & \textbf{0.68} \\
        \bottomrule
    \end{tabular}
\end{table}

\subsubsection{TSD neighbor Top-K smoothing Ablation}
\label{subsubsec:TSD_ablation}

To assess the sensitivity of the Token Splatting Decoder (TSD) to the neighbor Top-K smoothing $K$, we conducted an ablation study with varying $K$ values.
The results are summarized in \cref{tab:ablation_TSD}. 
As observed, setting $K=1$ (equivalent to hard decoding) leads to sharp gradients and suboptimal convergence, resulting in an Average F1 of 0.77.
Increasing $K$ to 10 provides effective label smoothing, yielding the best overall performance (Average F1: 0.78).
However, further increasing $K$ to 20 results in over-smoothing, which dilutes the distinctive spatial features of the masks and degrades performance to 0.76.
Therefore, we adopt $K=10$ as the optimal setting.

\begin{table}[t]
\centering
\caption{Ablation study on the TSD neighbor Top-K smoothing $K$. We report \textbf{F1} scores. $K=10$ yields the best trade-off between gradient smoothing and feature distinctiveness.}
\label{tab:ablation_TSD}
\begin{tabular}{l|cccccc|c}
\toprule
$K$ Value & CASIA1 & NIST & Coverage & Columbia & Glide & IMD & \textbf{Average} \\
\midrule
$K=1$ (Hard) & \textbf{0.87} & 0.56 & 0.74 & 0.88 & \textbf{0.80} & 0.75 & 0.77 \\
$K=5$        & 0.85 & 0.54 & 0.74 & 0.88 & 0.79 & 0.75 & 0.76 \\
\rowcolor{yellow!15}
$K=10$ (Ours) & 0.85 & \textbf{0.58} & \textbf{0.77} & \textbf{0.90} & \textbf{0.80} & \textbf{0.76} & \textbf{0.78} \\
$K=20$       & \textbf{0.87} & 0.55 & 0.73 & 0.88 & 0.76 & 0.76 & 0.76 \\
\bottomrule
\end{tabular}
\end{table}

\subsection{Robustness Study}

\begin{table}[t]
\centering
\caption{Robustness comparison (F1 scores) under various perturbations. Our method demonstrates superior stability across all degradations.}
\label{tab:robustness}
\resizebox{\textwidth}{!}{
\begin{tabular}{l|ccccccc}
\toprule
Method & Original & JPEG-80\% & JPEG-70\% & Gauss ($\sigma^2=10$) & Gauss ($\sigma^2=5$) & Resize 0.5 & Resize 0.75 \\
\midrule
FakeShield & 0.46 & 0.42 & 0.41 & 0.44 & 0.45 & 0.42 & 0.43 \\
SIDA & 0.57 & 0.54 & 0.54 & 0.51 & 0.54 & 0.54 & 0.53 \\
\rowcolor{yellow!15}
\textbf{ForensicsTok} & \textbf{0.79} & \textbf{0.68} & \textbf{0.66} & \textbf{0.71} & \textbf{0.73} & \textbf{0.67} & \textbf{0.73} \\
\bottomrule
\end{tabular}%
}
\end{table}

To evaluate robustness against real-world degradations \cite{wu2022robust, bayar2016deep}, we apply JPEG compression (quality levels of 80\% and 70\%), Gaussian 
noise (variance $\sigma^2=5$ and $10$), and resizing distortions (factor of 0.75 and 0.5) to test images across all datasets.
To keep the original resolution under resizing, we up-sample the images with corresponding scaling factor using bicubic interpolation.
We compare ForensicsTok with FakeShield~\cite{xu2025fakeshield} and SIDA~\cite{huang2025sida} by reporting average F1 scores under these perturbations. We did not re-run all expert baselines under every perturbation in this quick robustness study, so the table should be interpreted as an MLLM-baseline robustness comparison rather than a complete robustness ranking.
As shown in \cref{tab:robustness}, ForensicsTok consistently outperforms the baselines across all perturbations.
This robustness is consistent with the role of the Hierarchical Expert Fusion (HEF) module, which extracts multi-scale forensic features.
Notably, while the model maintains its lead, we observe a relatively larger performance drop under JPEG-70\% compression (F1 of 0.66).
This occurs because strong compression tends to blur critical fine-grained forensic clues.
This phenomenon suggests that ForensicsTok relies on fine-grained tampering traces in addition to high-level semantics, and also highlights a failure mode under strong compression.

\subsection{Computational Efficiency Analysis}
\label{subsubsec:efficiency}

\cref{tab:efficiency} presents a rigorous breakdown of computational costs.
Compared to the MLLM-based baseline FakeShield~\cite{xu2025fakeshield}, ForensicsTok demonstrates better resource efficiency within the MLLM category. By streamlining the architecture, we reduce total parameters by 62.5\% (from $\sim$22.7 B to $\sim$8.5 B) and obtain a 1.6$\times$ speedup in inference latency (from 2.26 s to 1.39 s), while improving F1 from 0.46 to 0.78 under our protocol.
However, ForensicsTok is still far heavier than SparseViT~\cite{su2025can}.
It should therefore be viewed as an MLLM-compatible localization framework rather than a replacement for lightweight experts in latency-critical deployments.

\begin{table}[tb]
\centering
\caption{Computational efficiency comparison. To ensure clarity, we detail the parameter breakdown for each component (\eg, MLLM backbone vs. HEF module). For FakeShield~\cite{xu2025fakeshield}, \textbf{DTE} refers to the \textbf{Domain Tag-guided Explainable Forgery Detection module} (DTE-FDM), and \textbf{MFLM} refers to the \textbf{Multi-modal Forgery Localization Module}.}
\label{tab:efficiency}
\resizebox{\textwidth}{!}{
\begin{tabular}{l|l|l|c|c|c}
\toprule
\textbf{Model} & \textbf{Encoder Params} & \textbf{Decoder Params} & \textbf{Total} & \textbf{Time} & \textbf{F1} \\
\midrule
SparseViT~\cite{su2025can} & 49.68 M & 0.66 M & $\sim$0.05 B & 0.04 s & 0.76 \\
FakeShield & 13.35 B (DTE-FDM) & 8.67 B (MFLM) + 641.09 M (SAM) & $\sim$22.66 B & 2.26 s & 0.35 \\
\midrule
\rowcolor{yellow!15}
ForensicsTok & \textbf{8.08 B (MLLM) + 99.27 M (HEF)} & \textbf{340.88 M (TSD)} & \textbf{$\sim$8.52 B} & \textbf{1.39 s} & \textbf{0.79} \\
\bottomrule
\end{tabular}%
}
\end{table}

\section{Conclusion}
This work proposes an MLLM-based IML framework, ForensicsTok, by reformulating the IMDL task as an autoregressive sequence generation task.
ForensicsTok combines a codebook-based mask-token interface with the TSD smoothing objective and the HEF module for injecting discriminative forensic features.
Experiments show large gains over previous MLLM-based localization baselines and competitive performance relative to strong forensic experts, with a clear efficiency trade-off compared with lightweight expert models.
In the future, we plan to investigate the explainable text output of ForensicsTok and evaluate the framework on broader realistic and AIGC manipulation benchmarks.

%
\clearpage
\bibliographystyle{unsrt}
\bibliography{main}
\end{document}